\documentclass[aps,pra,reprint,nofootinbib,longbibliography]{revtex4-2}

\usepackage[T1]{fontenc}
\usepackage[utf8]{inputenc}
\usepackage{graphicx}
\usepackage{amsmath,amssymb,amsfonts}
\usepackage{siunitx}
\usepackage{booktabs}
\usepackage{xcolor}
\usepackage[hidelinks]{hyperref}
\usepackage[nameinlink,noabbrev]{cleveref}

\newcommand{\gammaone}{\gamma_1}
\newcommand{\gammaphi}{\gamma_\phi}
\newcommand{\liouvillian}{\mathcal{L}}
\newcommand{\featurecount}{k}

\begin{document}

\title{A Spectral Identifiability Threshold for Dissipative Rate Recovery from Truncated Liouvillian Spectra}

\author{Yujun Ji}
\affiliation{Dulwich College Shanghai, Shanghai, China}

\author{Somyajit Chakraborty}
\email{chksomyajit@sjtu.edu.cn}
\affiliation{Department of Chemical Engineering, School of Chemistry and Chemical Engineering, Shanghai Jiao Tong University, Shanghai 200240, People's Republic of China}

\begin{abstract}
Open quantum systems lose energy and phase coherence through different
dissipative processes, but these processes can produce overlapping dynamical
signatures. The Liouvillian spectrum summarizes how such a system relaxes, yet
it is not obvious how much of that spectrum is needed to distinguish the
underlying dissipation rates. We study this question for amplitude damping and
dephasing in a six-qubit Lindblad model whose spectrum can be derived
analytically. We retain only the slowest non-steady spectral modes and ask how
many are required before each dissipative rate becomes recoverable. We show
that population modes contain no dephasing information, which creates a lower
bound of \(D=2^n\) retained modes for uniform dephasing identifiability in the
relevant rate regime. The measured recovery threshold reaches this bound at
\(n=4,5,6\), while \(n=3\) remains above it. At \(n=6\), least squares achieves
a mean joint absolute error of order \(10^{-9}\), compared with
\(4.355\times10^{-4}\) for four tabular learning methods. Robustness tests show
that this advantage weakens when the spectra are perturbed and when a
transverse field breaks the commuting structure. These results show that the
amount and structure of retained spectral information can determine whether
dissipative parameters are recoverable, independently of the estimator used.
The present conclusions apply to noise-free simulator spectra rather than
measurement-derived spectra.
\end{abstract}

\keywords{open quantum systems, Liouvillian spectra, identifiability, inverse problems, machine learning, dissipative dynamics}

\maketitle

\section{Introduction}
Quantum devices are never perfectly isolated from their surroundings, and these
environmental interactions lead to energy loss, loss of coherence, and other
forms of dissipative dynamics. In the Markovian regime, this evolution is
represented by a Lindblad generator \(\liouvillian\), which acts on density
operators and compactly encodes both Hamiltonian and dissipative contributions
\cite{Gorini1976,Lindblad1976,BreuerPetruccione2002}. The spectrum of this
generator is physically informative: nonzero real parts determine decay time
scales, imaginary parts encode oscillatory components, and the zero mode is
associated with steady-state structure. Dissipation can also be a useful
resource for state engineering and computation rather than only a source of
loss~\cite{Verstraete2009}. These properties make Liouvillian spectra a natural representation for inverse
problems that ask whether the physical rates governing dissipation can be recovered from the observed dynamical structure.

The primary motivation for this study is the identification of the physical
rates that control dissipation. In a real quantum device,
amplitude damping and dephasing rates determine coherence loss, gate quality,
and the reliability of downstream quantum-control protocols. Standard
identification strategies often use time-domain measurements, steady-state
constraints, tomography, or process-learning protocols. Representative recent
work has inferred open-system dynamics from steady states, learned
non-Markovian quantum dynamics from data, performed Lindblad tomography on
superconducting hardware, and studied learning guarantees for unknown quantum
processes~\cite{Bairey2020,Luchnikov2020,Samach2022,Huang2023Process}.
Hamiltonian and Liouvillian learning have also been used for verification of
digital quantum simulation~\cite{Pastori2022}, while quantum-tailored
machine-learning architectures have been proposed for superconducting-qubit
characterization~\cite{Genois2021}.

Despite these advances, a more basic representation question remains unresolved. Much of the current
identification literature asks how to recover generators, processes, or noise
parameters from experimentally accessible data: steady states, time traces,
local projective measurements, simulation-assisted measurement data,
randomized-benchmarking records, or quantum-jump trajectories
\cite{Bairey2020,Samach2022,Cemin2024LocalGenerators,WangLi2024SimulationAssisted,Zhang2025CorrelatedNoise,Radaelli2026JumpUnraveling}.
Another recent direction emphasizes structure-aware or graybox estimators that
embed physical assumptions, measurement-cost accounting, or ansatz selection
directly into the learning protocol
\cite{Youssry2024Graybox,Franca2024Hamiltonians,Olsacher2025LiouvillianLearning}.
A parallel line asks what is identifiable in principle from a given
experimental setup, classifying attainable knowledge for controllable systems
and reconstructing Hamiltonians from measurement time traces
\cite{Burgarth2012Identification,Zhang2014HamiltonianID}. Here the central question is whether the retained spectral representation
contains enough information to distinguish the rates at all. We therefore use
identifiability in an algebraic rather than statistical sense. We ask whether a
truncated sorted spectrum uniquely determines the rates, rather than with what
variance a particular estimator recovers them. Quantum Fisher information and
Cram\'er--Rao treatments of decoherence-rate estimation
\cite{BraunsteinCaves1994,Paris2009Estimation} therefore address a different
quantity from the threshold studied here.
These directions are essential for hardware-facing characterization, but they
do not isolate the complementary spectral-sufficiency question addressed here.
If a simulator or theoretical analysis already supplies a truncated Liouvillian
spectrum rather than a measurement record, how much of that spectrum must be
retained before the underlying dissipative rates become identifiable?
This question is also motivated by the role of low-lying Liouvillian
eigenvalues in open-system dynamics: modes with real parts closest to zero
govern the slowest relaxation processes, and separated low-lying sectors are
central to spectral descriptions of metastability and dissipative gaps
\cite{BreuerPetruccione2002,Macieszczak2016Metastability,Minganti2018Spectral}.

\begin{figure*}[t]
  \centering
  \includegraphics[width=\textwidth]{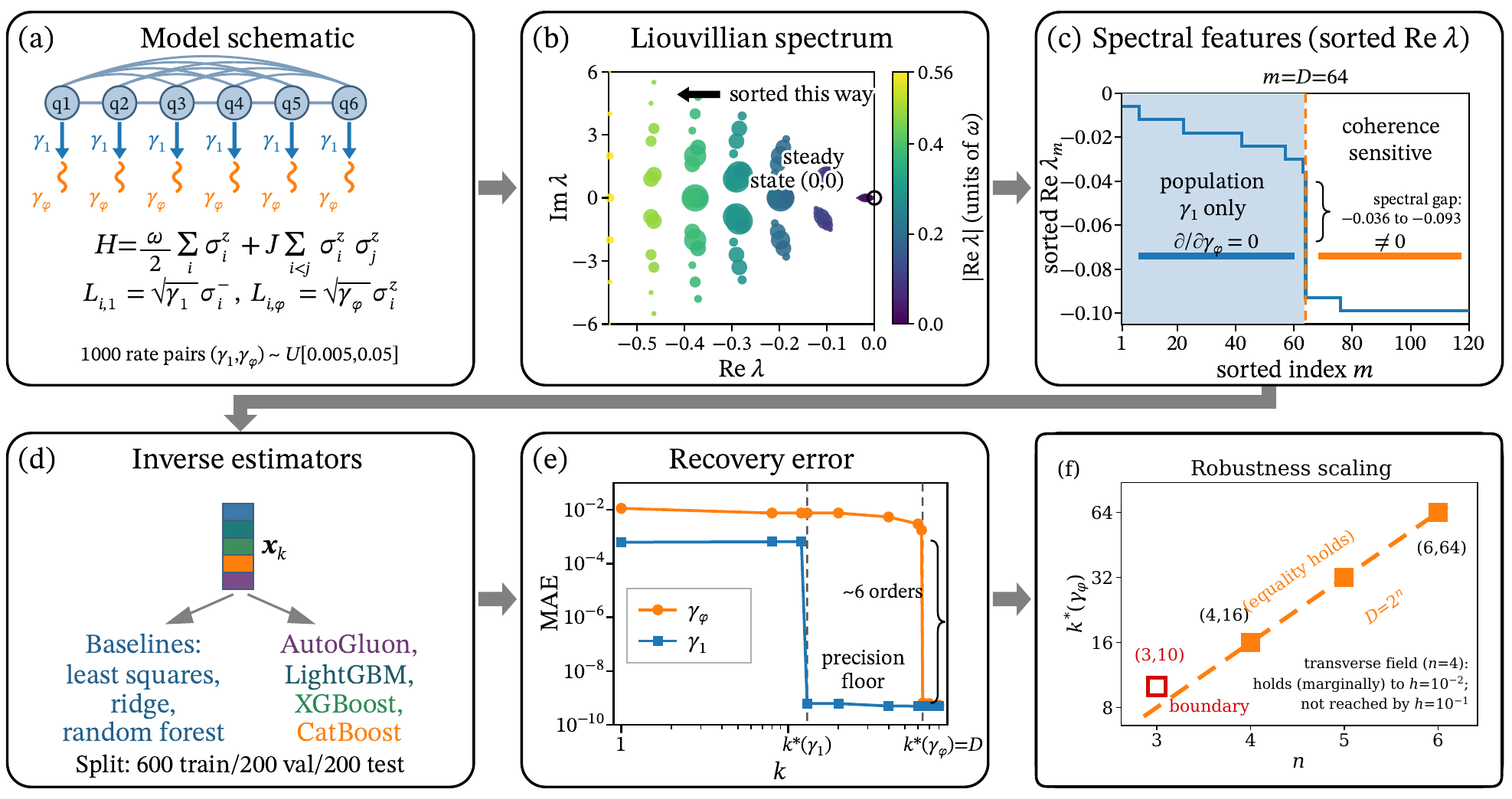}
  \caption{Workflow of the direct-spectrum identifiability benchmark.
(\textbf{a}) Six-qubit Lindblad model with local amplitude damping
\(\gammaone\) and dephasing \(\gammaphi\).
(\textbf{b}) Non-steady Liouvillian eigenvalues after removal of the steady mode.
(\textbf{c}) Eigenvalues are ordered by decreasing real part to form truncated
spectral inputs. In the populations-first regime, the first \(D-1\) non-steady
population modes are independent of \(\gammaphi\), and the first
dephasing-sensitive coherence appears at \(m=D\), yielding the identifiability
requirement \(\featurecount\ge D\).
(\textbf{d}) Linear and tabular estimators map the retained spectral features to
\((\gammaone,\gammaphi)\).
(\textbf{e}) The measured recovery thresholds occur at
\(\featurecount^\star=13\) for \(\gammaone\) and
\(\featurecount^\star=64=D\) for \(\gammaphi\) at \(n=6\).
(\textbf{f}) The dephasing threshold saturates the derived requirement at the
tested sizes \(n=4,5,6\), while \(n=3\) remains above the bound; the inset
summarizes the transverse-field robustness test at \(n=4\).}
  \label{fig:intro-inverse-problem}
\end{figure*}

To study this question under controlled conditions, we formulate a supervised
inverse benchmark. For each sample,
a synthetic six-qubit open quantum system is assigned an amplitude-damping
coefficient \(\gammaone\) and a dephasing Lindblad coefficient \(\gammaphi\).
The Liouvillian eigenvalues are computed, the steady-state eigenvalue is
removed, and the slowest \(\featurecount\) non-steady modes are converted into
real-valued tabular features. The inverse task is to recover the two coefficients
\((\gammaone,\gammaphi)\) from these truncated spectral features (\cref{fig:intro-inverse-problem}). The analysis first asks when this inverse map becomes recoverable with simple
linear regressors. We then compare AutoGluon, LightGBM, XGBoost, and CatBoost
on a matched grid
\(\featurecount \in \{80,160,240,320,400\}\), using three
train/validation/test trials per feature budget as a structured-data stress test
\cite{Erickson2020AutoGluon,Ke2017LightGBM,ChenGuestrin2016,Prokhorenkova2018CatBoost}.
Across the 111-dataset tabular benchmark of Shmuel et al., which ranks 20
model configurations, the
three strongest methods by average rank are AutoGluon, CatBoost, and
LightGBM; all three are included here, together with XGBoost as the most
widely used boosting implementation
\cite{Shmuel2025TabularBenchmark}.
A separate AutoGluon-only sweep over \(\featurecount=10,\ldots,70\) probes the
onset of useful information below the matched grid, and is treated as a
resolution study rather than a cross-model comparison.

The question we address is identifiability from a truncated, sorted and
mode-unlabelled slow Liouvillian spectrum. Given only the leading decay rates,
with the physical mode labels removed by sorting, what is the minimum spectral
depth required before each dissipative rate becomes uniformly recoverable? The
contribution is threefold. First, we identify the mechanism. This commuting family has a closed-form
spectrum that is affine in both rates, while its population sector is completely
independent of dephasing. This creates a lower bound on how much of the spectrum
must be retained before \(\gammaphi\) can become identifiable from the
representation. Second, we test when the measured recovery threshold saturates that bound.
Saturation occurs at \(n=4,5,6\), but not at \(n=3\), where the bound remains
valid without being attained. The measured equality also survives a
commutation-breaking transverse field up to \(h=10^{-2}\). Neither the
saturation nor this field scale follows from the analytical spectrum alone.
Both are measured numerically. Third, we measure how standard tabular learners perform on the same inverse map when the affine structure is not supplied explicitly, and we test how the
advantage of linear recovery changes under spectral perturbation,
using matched runs, target-wise metrics, and post hoc diagnostics such as
SHapley Additive exPlanations (SHAP) and permutation importance
\cite{LundbergLee2017,Lundberg2020TreeSHAP}. Throughout, sorted-coordinate
attributions are treated as predictive diagnostics rather than physical
eigenmode identities.

\section{Results}

\subsection{A closed-form spectrum and where each rate becomes identifiable}
Every term of the Hamiltonian and every dephasing operator is diagonal in the
computational basis, while each damping operator strictly lowers the
excitation number. The vectorized generator is therefore triangular in the
ordered computational-pair basis, and its eigenvalues are its diagonal
entries: the model belongs to a known exactly solvable class of Lindblad
master equations without gain~\cite{Torres2014ClosedForm}. It is distinct from
the quadratic or quasi-free class solved by third quantization
\cite{Prosen2008ThirdQuantization}, which requires Lindblad operators linear in
the mode variables; here solvability follows instead from simultaneous
diagonality of \(H\) and the dephasing channel. Indexing the triangular Liouvillian representation by ordered computational-basis pairs \((i,j)\)
with diagonal energies \(E_i\), excitation numbers \(n_i\), and Hamming
distance \(d_H(i,j)\),
\begin{equation}
  \lambda_{ij}
  =
  -\mathrm{i}\left(E_i-E_j\right)
  -
  \frac{\gammaone}{2}\left(n_i+n_j\right)
  -
  2\gammaphi\, d_H(i,j),
  \label{eq:closed-form}
\end{equation}
so every eigenvalue is affine in \((\gammaone,\gammaphi)\) (see Methods for
the derivation and the excitation-number convention). We verified
\cref{eq:closed-form} entrywise against the numerically constructed generator
at \(n=4\) over 25 protocol rate pairs, with maximum absolute deviation
\(1.67\times10^{-16}\) and exactly vanishing upper-triangular residual, and at the sorted-multiset
level for \(n=3,\ldots,6\). The near-linear identifiability below is therefore structural to this
commuting family.

The simple-baseline diagnostics, fitted on the 600-row benchmark-arm partition, show where that identifiability sets in: the map
becomes essentially invertible by least squares once the retained slow sector
is large enough, and the two coefficients reach that point at different
budgets. Least squares
recovers \(\gammaone\) to a mean absolute error (MAE) of
\(5.94\times10^{-10}\) already at
\(\featurecount=20\), whereas its \(\gammaphi\) error is still
\(6.38\times10^{-3}\) at that budget and improves only gradually, to
\(2.65\times10^{-3}\), by \(\featurecount=60\). Between \(\featurecount=60\)
and \(\featurecount=70\) the \(\gammaphi\) error falls to
\(9.08\times10^{-10}\), so the mean joint MAE drops from
\(1.32\times10^{-3}\) to \(8.56\times10^{-10}\), more than six orders of
magnitude across a single step of the budget grid. It remains in the same \(\approx10^{-9}\) quantization-limited regime at \(\featurecount=80\)
\((8.52\times10^{-10})\) and at every larger budget tested, ranging over
\(7.4\times10^{-10}\) to \(3.0\times10^{-9}\). A finer scan on the separate 800/200 threshold-scan partition resolves both onsets. Fitting ordinary least
squares on an intercept and the first \(\featurecount\) real coordinates alone,
for every budget \(\featurecount=1,\ldots,80\), the smallest budget at which
the held-out MAE falls below \(10^{-7}\) is \(\featurecount^\star=13\) for
\(\gammaone\) and \(\featurecount^\star=64\) for \(\gammaphi\); both values are
identical for split seeds 42, 43, and 44. Because those three seeds partition
the same 1000 rate pairs, this is a stability check against the particular
partition rather than replication on independent data; the mechanism derived
below, not the seed agreement, is what forces
\(\featurecount^\star(\gammaphi)\) to be no smaller than \(D\) (see Methods). The
\(\gammaphi\) transition is confined to one coordinate: for split seed 42 the
held-out MAE is \(1.77\times10^{-3}\) at \(\featurecount=63\) and
\(6.40\times10^{-10}\) at \(\featurecount=64\), and seeds 43 and 44 collapse
at the same coordinate, from \(2.11\times10^{-3}\) and \(2.22\times10^{-3}\)
to below \(10^{-9}\). The coarse benchmark grid brackets the two onsets
rather than resolving them; the finer scan pins
\(\featurecount^\star(\gammaphi)=64\), which equals the Hilbert-space
dimension \(D=2^n=64\). \Cref{eq:closed-form} shows why \(\featurecount^\star(\gammaphi)\) cannot
fall below \(D\); that it lands exactly on \(D\) at these sizes is measured,
not implied by the algebra.
Ridge regression reproduces the same behaviour, with mean joint MAE
\(4.48\times10^{-6}\) at \(\featurecount=70\) and \(1.34\times10^{-5}\)
at \(\featurecount=80\); the effect therefore follows from the retained
spectral sector rather than from one unregularized solver. The ridge value is not a limiting value. Its penalty is selected by leave-one-out cross-validation, and the
selected \(\alpha\) falls from \(4.1\)--\(9.3\times10^{-4}\) at
\(\featurecount=80\) to \(5.9\times10^{-7}\)--\(3.0\times10^{-6}\) at
\(\featurecount=320\) and 400, where the design is underdetermined and ridge collapses onto the
least-squares solution \((1.9\times10^{-9}\) at \(\featurecount=320)\).
Ridge values at a single budget should be read together with the selected
penalty.

\subsection{A derived lower bound on the retained window}
The dephasing threshold follows from the population sector of
\cref{eq:closed-form}. Population modes (\(i=j\), \(d_H=0\)) have
\(\operatorname{Re}\lambda=-\gammaone n_i\) and are blind to \(\gammaphi\);
there are \(D\) of them, and the non-steady filter removes exactly the one
with \(n_i=0\), leaving \(D-1\) dephasing-blind non-steady modes. Whether
these blind modes fill the head of the sorted spectrum depends on the rate
ratio: every population precedes every coherence exactly when
\(\gammaphi/\gammaone>(2n-1)/4\), obtained by equating the fastest population
rate \(\gammaone n\) with the slowest coherence rate
\(\gammaone/2+2\gammaphi\). The sampled rectangle \([0.005,0.05]^2\) contains
pairs on both sides of this line, so the position of the first
\(\gammaphi\)-informative coordinate varies across samples --- from 1 to \(D\) at every tested system size, with 120 of the 1000 pairs
attaining \(D\) at \(n=6\). A
single linear map fitted across the whole sample must therefore retain
\(\featurecount\ge D\) coordinates before \(\gammaphi\) is identifiable in its worst-case samples. The bound is a statement of non-injectivity, not merely of
missing labels: fixing \(\gammaone\) and taking any two distinct
\(\gammaphi\) values in the populations-first regime leaves the first
\(D-1\) sorted non-steady coordinates numerically identical, because those
coordinates are all populations and populations do not depend on
\(\gammaphi\). The two parameter values are then indistinguishable from any
prefix with \(\featurecount<D\), and first differ at coordinate \(D\).
Hence the requirement \(\featurecount\ge D\) is a derived lower bound for any estimator using sorted prefixes, for any ensemble containing two distinct \(\gammaphi\) at a common \(\gammaone\) in this regime, while the measured recovery threshold \(\featurecount^\star(\gammaphi)\) is the smallest budget at which the held-out error crosses the reported cutoff. The bound guarantees the presence of dephasing information at
\(\featurecount=D\), not the solvability or conditioning of the resulting
linear system; that the measured threshold saturates the bound is the
empirical content of the scan. We deliberately estimate the rates by
least squares on the retained coordinates rather than by inverting
\cref{eq:closed-form} directly. A direct inversion would presuppose the mode
labelling \((i,j)\), which the sorted, degenerate spectral coordinates do not
supply; the regression setting is also the one in which the tabular
comparison below is meaningful.

The same crossover controls the \(\gammaone\) onset. The slowest non-steady
mode is the single-excitation population with rate \(\gammaone\) unless a
coherence is slower, which happens exactly when \(\gammaphi<\gammaone/4\); 60
of the 1000 sampled pairs (\(6.0\%\)) fall in this regime, consistent with
the \(5.56\%\) area fraction that this condition occupies in the sampled
rectangle. The leading sorted
coordinate \(a_1\) therefore equals \(-\gammaone\) on most, but not all,
samples, which is why the one-coordinate estimator
\(-\texttt{eig\_real\_000}\) is not an exact identity (see Methods) and why
\(\featurecount^\star(\gammaone)\) exceeds one: the fit must separate the two
slowest-mode identities. A companion bound follows from the same construction.
In the coherences-first regime the \(2n\) slowest non-steady modes are the
\((n_i,n_j)=(0,1)\) coherences, degenerate at
\(-(\gammaone/2+2\gammaphi)\); a prefix confined to them is rank-one in the
two rates, and the first population coordinate enters at sorted index
\(2n+1\), which we verified exactly at \(n=3,\ldots,7\). Hence \(\featurecount\ge 2n+1\) is the companion requirement, attained by
the measured threshold at \(n=4,5,6\) (9, 11, 13). At \(n=3\) the measured
\(\gammaone\) threshold is \(\featurecount^\star=10\), above \(2n+1=7\): the
requirement holds there as everywhere, but is not attained, mirroring the
\(\gammaphi\) threshold at that size. Both
thresholds coincide at 10 for \(n=3\), and \(n=3\) is the only size tested at
which neither is attained.
 The transition is a single-coordinate step, not a gradual decline: at
\(n=6\) the held-out \(\gammaone\) MAE sits at \(6.3\)--\(6.5\times10^{-4}\)
through \(\featurecount=12\) and falls to \(6.13\times10^{-10}\) at
\(\featurecount=13\) (split seed 42).

\subsection{System size and the role of commutation}
Repeating the generation and scan protocol at \(n=3\), 4 and 5 with identical
physics tests the bound across system size, as summarized in panel f of \cref{fig:intro-inverse-problem}. The dephasing threshold equals \(D=2^n\) at \(n=4,5,6\), identically across split seeds, while the required fraction of the non-steady spectrum shrinks from
\(15.9\%\) to \(1.6\%\) between \(n=3\) and \(n=6\). At \(n=3\) the measured threshold is \(\featurecount^\star=10\), above
\(D=8\); the bound holds there as everywhere, but is not attained. We
therefore report the equality \(\featurecount^\star(\gammaphi)=D\) as an
observation at the sizes tested, \(n=4,5,6\), rather than as a general result for all \(n\). The derived statement is the requirement \(\featurecount\ge D\); whether a least-squares map attains
it at a given \(n\) is a separate, measured question.

The commuting structure, not merely the model family, is what carries the
threshold.
Adding a uniform transverse field \(h\sum_i\sigma_i^x\) at \(n=4\) and
repeating the scan (see Methods) shows that
\(\featurecount^\star(\gammaphi)=D=16\) survives unchanged, on all three
split seeds, through \(h=10^{-4}\), \(10^{-3}\), and \(10^{-2}\), while the
\(\gammaone\) threshold drifts from 9 to 14 at \(h=10^{-2}\). That row sits
close to the criterion: its held-out \(\gammaphi\) MAE at
\(\featurecount=D\) is \(8.6\times10^{-8}\), within \(20\%\) of the
\(10^{-7}\) cutoff, against a margin of more than two orders at \(h=0\), so
the verdict for \(h=10^{-2}\) should be read as marginal rather than clear; by \(h=10^{-1}\)
neither threshold is reached within the scanned budget \(\featurecount\le32\) (\cref{tab:transverse}). On this grid the loss of the equality is bracketed only at decade resolution, between \(h=10^{-2}\) and
\(10^{-1}\). The measured deviation of the sorted spectrum from the \(h=0\)
closed form grows by close to two decades per decade of \(h\) above the numerical resolution limit (\(1.1\times10^{-6}\), \(1.1\times10^{-4}\), and
\(1.0\times10^{-2}\) at \(h=10^{-3}\), \(10^{-2}\), and \(10^{-1}\)),
compatible with approximately quadratic scaling on this coarse grid. Because the uniform
field preserves site-permutation symmetry, symmetry-sector degeneracies
persist at \(h>0\); the sorting convention was left unchanged.

\begin{table*}[!htbp]
\centering
\caption{Transverse-field robustness of the threshold bound at \(n=4\)
(\(D=16\)). Thresholds are identical across split seeds 42, 43, and 44 at
every \(h\); ``\(>32\)'' records right-censoring at the scan bound. The last
column is the maximum absolute deviation of the sorted non-steady
\(\operatorname{Re}\lambda\) multiset from the \(h=0\) closed form over 25
protocol rate pairs. This is an eigensolver-level comparison of sorted
multisets; the \(h=0\) entry is its numerical resolution limit and is not the entrywise
float64 residual of \(1.67\times10^{-16}\) reported in the text}
\label{tab:transverse}
\begin{tabular}{ccccc}
\toprule
\(h\) & \(\featurecount^\star(\gammaone)\) & \(\featurecount^\star(\gammaphi)\)
& verdict & max deviation \\
\midrule
0          & 9       & 16      & \(\featurecount^\star(\gammaphi)=D\) & \(1.4\times10^{-8}\) \\
\(10^{-4}\) & 9       & 16      & \(\featurecount^\star(\gammaphi)=D\) & \(2.1\times10^{-8}\) \\
\(10^{-3}\) & 9       & 16      & \(\featurecount^\star(\gammaphi)=D\) & \(1.1\times10^{-6}\) \\
\(10^{-2}\) & 14      & 16      & \(\featurecount^\star(\gammaphi)=D\) (marginal) & \(1.1\times10^{-4}\) \\
\(10^{-1}\) & \(>32\) & \(>32\) & not reached & \(1.0\times10^{-2}\) \\
\(5\times10^{-1}\) & \(>32\) & \(>32\) & not reached & \(1.1\times10^{-1}\) \\
\bottomrule
\end{tabular}
\end{table*}

\subsection{Structure-agnostic learners on the same map}
Having established where each rate becomes identifiable by a linear estimator, we
now ask what standard tabular learners achieve on the same features when they
are not told the affine structure. The comparison uses the matched
high-resolution benchmark: 60 saved runs formed by four model families, five
feature budgets \(\featurecount=80,160,240,320,400\), and three trials per
budget (see Methods). Each run evaluates 200 held-out test samples, giving 15
complete \((\featurecount,\mathrm{trial})\) blocked units shared by all four
models.

\begin{figure*}[!htbp]
  \centering
  \includegraphics[width=\textwidth]{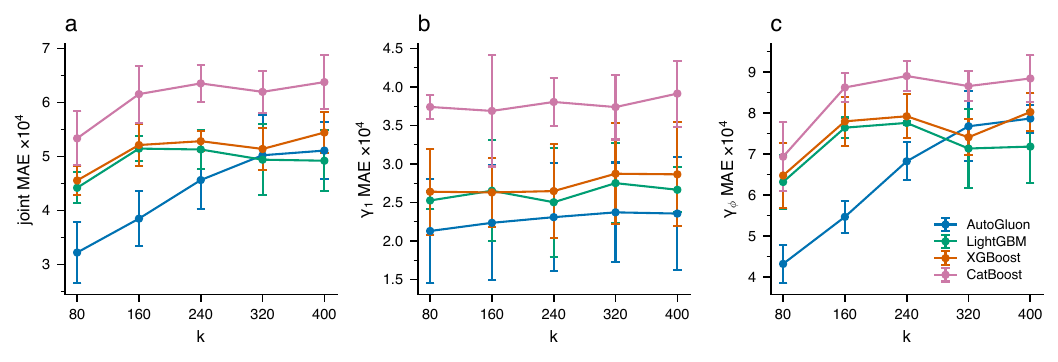}
  \caption{Matched comparison curves over \(\featurecount=80,\ldots,400\).
  (\textbf{a}) Joint MAE, (\textbf{b}) \(\gammaone\) MAE, and
  (\textbf{c}) \(\gammaphi\) MAE. Markers show the mean over three trials and
  error bars one standard deviation across them. The three trials are
  overlapping partitions of the same 1000 rate pairs, not independent
  datasets, so the bars are descriptive spread rather than sampling error}
  \label{fig:matched-model-comparison}
\end{figure*}

\Cref{fig:matched-model-comparison} shows that all four families plateau in the
same narrow band. Mean joint MAE is \(4.355\times10^{-4}\) for AutoGluon,
with a descriptive 95\% bootstrap interval of \([3.905,4.789]\times10^{-4}\),
followed by LightGBM at \(4.914\times10^{-4}\), XGBoost at
\(5.129\times10^{-4}\) and CatBoost at \(6.086\times10^{-4}\); the best single
model-budget point is AutoGluon at \(\featurecount=80\), where the mean joint
MAE is \(3.223\times10^{-4}\). We do not read this ordering as a ranking of
the libraries: the families span a factor of \(1.4\), and the ordering already
changes with the objective, since LightGBM has the highest average
\(\gammaone\) \(R^2\), \(0.9968\). What the benchmark
establishes is the plateau itself. Three of the four families were tuned over
multi-dimensional search spaces and agree within \(24\%\) of one another, with
AutoGluon --- run at a fixed quality preset (see Methods) --- falling in the
same band; all four sit five to six orders of magnitude above what least
squares achieves on identical inputs.

That gap is a hypothesis-class mismatch rather than a penalty for withholding
\cref{eq:closed-form}: tree ensembles are piecewise constant and cannot
represent an affine map exactly, whereas a linear estimator matches it without
being told the structure. It is a protocol-specific reference point, not a claim about the ceiling of
automated machine learning. The complete model-by-budget metrics are included in the archived analysis tables~\cite{ji_2026_22160077}.

The joint error averages two parameters the models do not recover equally
well. Split by target, every family places the larger absolute error on
\(\gammaphi\) (\cref{fig:matched-model-comparison}b,c): for AutoGluon the mean
\(\gammaphi\) MAE is \(6.43\times10^{-4}\) against \(2.28\times10^{-4}\) for
\(\gammaone\), a ratio of approximately \(2.8\). This is what
\cref{eq:closed-form} predicts, since the dephasing signal enters only once the
retained window reaches \(D\). It does not indicate that dephasing regression
fails here: mean \(R^2\) exceeds \(0.995\) for \(\gammaone\) and \(0.990\) for
\(\gammaphi\) in every family. We therefore track \(\gammaphi\) rather than the joint error
through the resolution comparisons that follow.

Adding spectral coordinates beyond the threshold does not help. All four
families reach their best aggregated joint MAE at the smallest matched budget,
\(\featurecount=80\) (\cref{fig:matched-model-comparison}a), with AutoGluon
ahead at \(\featurecount=80,160,240\) and LightGBM at
\(\featurecount=320,400\); for the present distribution and training protocol
the slowest 80 non-steady modes of this six-qubit generator already carry the
useful rate information. \Cref{eq:closed-form} explains why: every real part
is an integer combination of \(n_i+n_j\) and \(d_H(i,j)\), so the retained
real coordinates span only 14 to 27 dimensions across the whole grid while the
nominal predictor count grows from 160 to 800. Deeper coordinates add nominal
dimension without target-relevant rank. At \(\featurecount=320\) and 400 the
design is underdetermined \((p=640\) and 800 against 600 training rows\()\)
and the linear fits are minimum-norm solutions. Extending the AutoGluon arm downwards under the same
60-evaluation budget traces the approach to that
plateau: \(\gammaphi\) MAE falls by \(92.41\%\) between \(\featurecount=10\)
and \(70\), joint MAE by \(89.09\%\), and the \(\gammaphi/\gammaone\) ratio
contracts from \(27.33\) to \(2.04\). Comparing \(\featurecount=70\) with
\(\featurecount=80\) at matched budget gives a further \(2.23\%\) decrease in
mean joint MAE, from \(3.296\times10^{-4}\) to \(3.223\times10^{-4}\), and a
\(\gammaphi\) MAE moving from \(4.422\times10^{-4}\) to \(4.315\times10^{-4}\);
with three paired trials this is not resolved as a sharp transition
\((p=0.533)\). We therefore read the curve as improving rapidly through the
low-\(\featurecount\) regime, flattening near
\(\featurecount=70\)--\(80\), and degrading thereafter, joint MAE rising by
\(58.6\%\) from \(\featurecount=80\) to \(\featurecount=400\), without
assigning a precise optimum from three overlapping splits.

Two further checks bound how much weight the ordering can carry. Directional
win counts over the same 15 blocks give AutoGluon over LightGBM on 10 of 15,
over XGBoost on 13 of 15, and over CatBoost on all 15. The blocks are not
independent: the budgets are nested views of one rate-pair set, the splits
overlap, and the budget main effect accounts for most of the variance in the
paired differences, so the effective sample size is closer to the five budgets
than to fifteen. Aggregating by budget leaves only the CatBoost comparisons
separated. We therefore report that CatBoost is consistently the weakest of the
four and treat the ordering among AutoGluon, LightGBM and XGBoost as unresolved
at this design's effective sample size. Paired tests, corrections, and effect sizes are included in the archived analysis tables~\cite{ji_2026_22160077} and are descriptive rather than confirmatory (see Methods). Residuals at the AutoGluon \(\featurecount=80\)
operating point are accurate in aggregate but heavy-tailed: D'Agostino--Pearson
statistics are 571.4 for \(\gammaone\) \((p=8.35\times10^{-125}\), skewness
\(-3.90\), excess kurtosis 42.73\()\) and 229.0 for \(\gammaphi\)
\((p=1.86\times10^{-50}\), skewness 1.31, excess kurtosis 10.71\()\). Since the
pooled residual set repeats target pairs across overlapping partitions, these
\(p\)-values are descriptive and the shape statistics are the informative
quantities. Residual distribution statistics and descriptive bootstrap summaries are included in the archived analysis tables~\cite{ji_2026_22160077}.

\subsection{What the sorted coordinates carry}

\begin{figure*}[!htbp]
  \centering
  \includegraphics[width=0.95\textwidth]{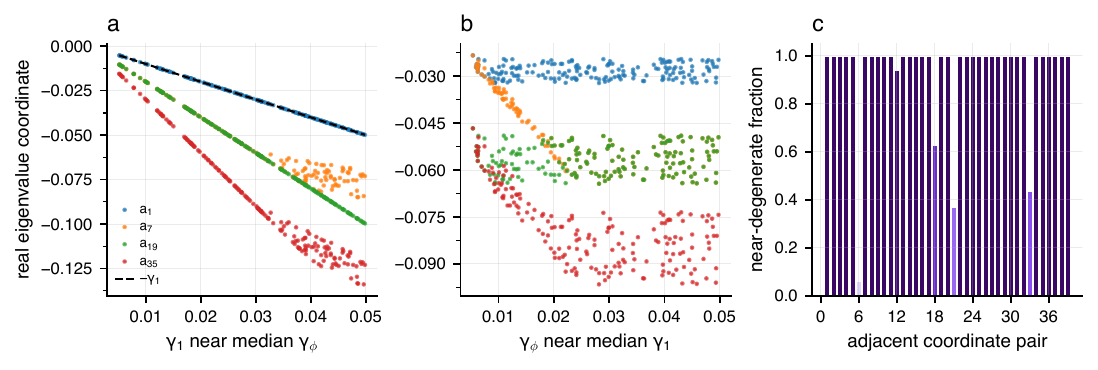}
  \caption{Low-lying Liouvillian coordinate diagnostics
  from the \(k=80\) dataset. (\textbf{a},\textbf{b}) Selected sorted real
  coordinates along narrow rate slices; (\textbf{c}) the fraction of
  samples with nearly degenerate adjacent sorted real coordinates. The
  degeneracy pattern is a rank-stability warning: sorted coordinates are useful
  predictive features, but should not be interpreted as globally tracked
  physical eigenmode identities without additional eigenvector tracking}
  \label{fig:nsr-eigen-trajectories}
\end{figure*}

\begin{figure*}[!htbp]
  \centering
  \includegraphics[width=0.95\textwidth]{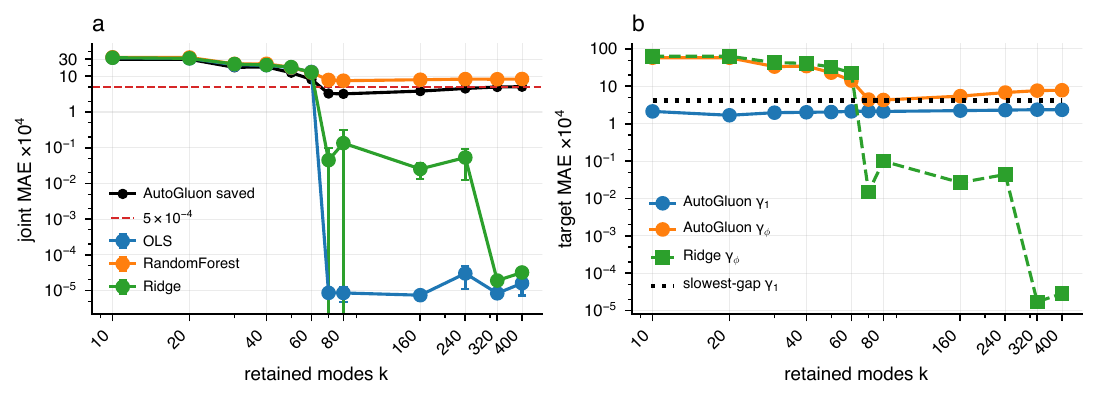}
  \caption{Analytical and simple-baseline diagnostics.
  (\textbf{a}) Joint MAE for ordinary least squares, ridge regression, and
  random forests fit on the same nested spectral CSVs used by the main
  benchmark, with the saved AutoGluon curve overlaid from the AutoGluon
  resolution summary. (\textbf{b}) Target-wise MAE for the saved AutoGluon
  runs, the ridge \(\gammaphi\) baseline, and the one-coordinate slowest-gap
  \(\gammaone\) estimator \(-\texttt{eig\_real\_000}\).
  The near-exact least-squares performance at larger \(k\) confirms the affine
  structure of \cref{eq:closed-form} for this fixed,
  computational-basis-diagonal generator. Thus the
  AutoGluon point should be read as the lowest-error point within the saved
  four-family boosted-tree/AutoML benchmark, not as a claim that AutoML is the
  best estimator for this inverse problem}
  \label{fig:nsr-baselines}
\end{figure*}

\begin{figure*}[!htbp]
  \centering
  \includegraphics[width=0.95\textwidth]{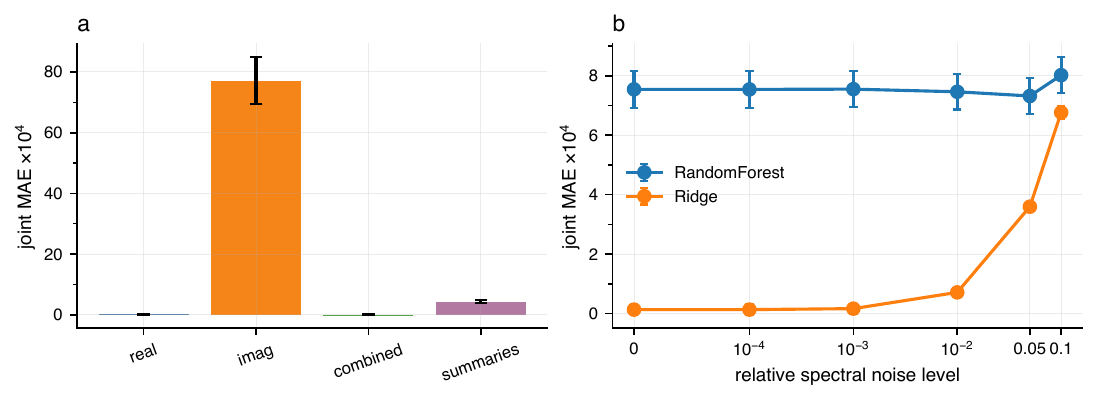}
  \caption{Real/imaginary feature ablation (\textbf{a}, ridge) and
  clean-train/noisy-test spectral robustness (\textbf{b}) for simple baselines
  at \(k=80\).
  Real parts carry nearly all of the target information for this fixed
  Hamiltonian family, while imaginary parts alone are weak --- as expected,
  since \(\operatorname{Im}\lambda_{ij}=-(E_i-E_j)\) in \cref{eq:closed-form}
  is independent of both rates. The noise panel
  (\textbf{b}) perturbs test spectral coordinates after fitting clean models;
  it is a diagnostic of feature sensitivity, not a retraining of the saved
  AutoGluon benchmark}
  \label{fig:nsr-ablation-noise}
\end{figure*}

\begin{figure*}[!htbp]
  \centering
  \includegraphics[width=0.95\textwidth]{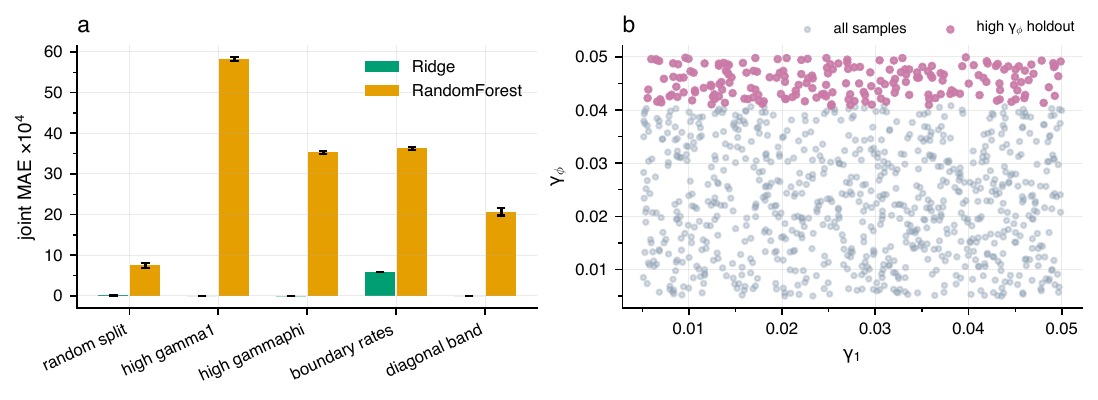}
  \caption{Unseen-rate-region holdout diagnostics on
  the \(k=80\) dataset. (\textbf{a}) Joint MAE across the four deterministic
  holdout schemes, together with the random-split reference;
  (\textbf{b}) example high-\(\gammaphi\) holdout region in the rate plane.
  Ridge regression extrapolates near-exactly onto the high-rate
  (\emph{high gamma1}) and diagonal (\emph{diagonal band})
  holdouts because the present spectral map is close to linear, but degrades
  to \(5.8\times10^{-4}\) on the two-sided boundary-margin region
  (\emph{boundary rates}); the random
  forest degrades on all four schemes and is the weaker extrapolator
  throughout. Ridge is more accurate on the high-rate and diagonal holdouts
  \((4.1\times10^{-8}\) and \(4.3\times10^{-8})\) than on the random split
  \((1.3\times10^{-5})\), whose value is inflated by one split seed and by the
  penalty cross-validation selects there; it is not a limiting value. This figure tests rate-plane generalization
  only, within the fixed commuting generator; commutation-breaking is assessed
  separately by the transverse-field scan reported above
  (\cref{tab:transverse})}
  \label{fig:nsr-ood-holdout}
\end{figure*}

The additional diagnostic suite adds six figure concepts: spectral trajectories
(\cref{fig:nsr-eigen-trajectories}), simple baselines
(\cref{fig:nsr-baselines}), target-wise benchmark confidence intervals, error
maps, feature ablations with spectral-noise perturbations
(\cref{fig:nsr-ablation-noise}), and an unseen-rate-region holdout
(\cref{fig:nsr-ood-holdout}). A validation audit confirms
that all \(k=10,\ldots,400\) CSV files use the same 1000 rate pairs in the same
order, and that recomputed prediction metrics agree with the dashboard records
to within \(5.449\times10^{-8}\). Target-wise and joint-MAE bootstrap intervals are included in the archived analysis tables~\cite{ji_2026_22160077}.

Two of these diagnostics bear directly on the closed-form structure.
\Cref{fig:nsr-eigen-trajectories}c shows that nearly every adjacent pair among
the first 40 sorted real coordinates is near-degenerate in a large fraction of
samples, which is expected from \cref{eq:closed-form}: eigenvalues depend only
on \(n_i+n_j\) and \(d_H(i,j)\), so many distinct modes share a real part and
sorted coordinates cannot be read as tracked eigenmodes.
\Cref{fig:nsr-ablation-noise}a shows that real coordinates alone recover the
targets almost as well as the full feature set while imaginary coordinates
alone are weak. For a fixed physical mode \cref{eq:closed-form} makes
\(\operatorname{Im}\lambda_{ij}=-(E_i-E_j)\) independent of both rates, but
the sorted sequence is not rate-independent: the rate-dependent ordering of
real parts determines which mode occupies each index, so imaginary coordinates
retain indirect rate information, making imaginary-only prediction weak rather
than exactly uninformative.

The advantage the affine structure confers has a finite noise budget.
Perturbing test coordinates with zero-mean Gaussian noise of relative scale
\(\eta\) after fitting on clean data (\cref{fig:nsr-ablation-noise}b), the
ridge joint MAE rises from \(1.34\times10^{-5}\) at \(\eta=0\) to
\(7.15\times10^{-5}\), \(3.60\times10^{-4}\) and \(6.76\times10^{-4}\)
at \(\eta=10^{-2}\), \(5\times10^{-2}\) and \(10^{-1}\), while the
random forest stays near \(7.5\times10^{-4}\). The linear advantage contracts
from \(56.2\) to \(10.4\), \(2.04\) and \(1.19\) across the same
sequence. The noise-free ratio is the least stable of these: per split seed it
ranges from \(20.9\) to \(390.1\), because the clean ridge error is itself in the quantization-limited regime. From \(\eta=10^{-2}\) onwards the ratio is tight
across seeds (\(9.2\)--\(11.7\), \(1.9\)--\(2.2\), \(1.1\)--\(1.3\)),
so the contraction is robust even though its starting point is not. The linear advantage measured here is therefore a property of near-exact spectra: it is already reduced to about one order of magnitude at \(\eta=10^{-2}\) and is gone by \(\eta=10^{-1}\). Because this scan compares ridge against a random forest, it bounds the persistence of the linear advantage under perturbation rather than tracking the five-to-six-order least-squares separation reported above; it is a clean-train/noisy-test diagnostic, not a retraining of the matched benchmark.

The most important diagnostic is that the inverse map of this fixed, commuting
generator is close to linearly identifiable from retained real spectral
coordinates. In
\cref{fig:nsr-baselines}, ordinary least squares and ridge regression strongly
outperform the saved AutoML/boosted-tree runs once the retained spectral sector
is sufficiently large. This does not invalidate the tabular benchmark, but it
changes its interpretation: the four-family comparison measures how standard
nonlinear tabular learners behave on a nearly invertible direct-spectrum task,
whereas the physical conclusion is that this commuting Liouvillian family
contains an almost linear spectral signature of \((\gammaone,\gammaphi)\).

For post hoc tree-based diagnostics we use the loadable XGBoost
\(\featurecount=80\) models as transparent boosted-tree baselines. This choice
does not change the four-family benchmark ranking, but it limits the
interpretation claim to the learned coordinates of one fitted regressor. The
SHAP beeswarm plots indicate that a small subset of sorted Liouvillian
coordinates has disproportionate influence on the XGBoost predictions. These
coordinates are operational features, not globally tracked physical eigenmode
identities. The \(\gammaphi\) beeswarm is emphasized because \(\gammaphi\) is
the harder and more representation-sensitive target; the companion
\(\gammaone\) beeswarm checks whether the easier target uses the same
coordinates or a distinct subset. The interpretability protocol and associated analysis scripts are included in the archived code release~\cite{ji_2026_22160077}.
These analyses audit one fitted model rather than establish the central
identifiability claim.

The gap between AutoGluon and the individual boosted-tree baselines is modest
but consistent under the mean joint-MAE objective. AutoGluon benefits from
automated model selection and ensembling, while LightGBM remains competitive
and overtakes AutoGluon in mean joint MAE at \(\featurecount=320\) and
\(\featurecount=400\). The saved search traces record validation-search trajectories, not per-epoch
training loss, and serve as reproducibility diagnostics.

Spectral features are not independent coordinates. Because sorted spectral
coordinates are frequently correlated and degenerate, individual feature
attributions should not be interpreted as isolated physical mechanisms.
The Spearman-correlation and clustering protocol used for this diagnostic is included in the archived analysis code~\cite{ji_2026_22160077}. Permutation
importance provides a complementary model-agnostic check. Because
sorted eigenvalue coordinates can change rank near degeneracies or crossings,
future measurement-facing versions of this benchmark should also test
permutation-invariant spectral summaries such as moments, empirical spectral
density features, kernel distances, or matched sets of eigenvalues rather than
relying only on coordinate-wise feature importance.

\section{Discussion}

The benchmark sits between two established directions in open-quantum-system
learning. On one side are tomography, process-learning, and steady-state
identification methods, in which the learner receives measurement data and
must infer a generator or its parameters. On the other are representation
studies asking which descriptors of a known generator carry the most useful
information. This work follows the second path: the input is the direct
Liouvillian spectrum, so every claim concerns a controlled spectral inverse
problem rather than experimental recovery from finite measurement records.

Within that scope the central result is the relationship between the derived identifiability requirement \(\featurecount\ge D\) and the measured recovery threshold \(\featurecount^\star(\gammaphi)=D\) for this commuting Liouvillian family, and it has two parts that should be kept distinct. The requirement \(\featurecount\ge D\) is derived: the commuting generator has
a closed-form spectrum affine in the two rates, its population sector is blind
to dephasing, and a sorted prefix shorter than \(D\) is therefore
non-injective in \(\gammaphi\). The bound is conditional on the sampled rate
ensemble: it binds because \([0.005,0.05]^2\) contains pairs with
\(\gammaphi/\gammaone>(2n-1)/4\), and an ensemble confined below that line
would place the threshold at \(O(n)\) instead. That the bound is
\emph{attained} is measured,
and it is not implied by the algebra. It is also a statement about a
mode-unlabelled representation: sorting discards the pair labels \((i,j)\), so
knowing \cref{eq:closed-form} does not let an estimator invert it, and the
constraint applies to any method reading sorted prefixes rather than only to
the ones tested here. The same mechanism explains the target
asymmetry: \(\gammaphi\) is the harder target precisely because
\cref{eq:closed-form} delays its information to coordinate \(D\) in the
worst-case samples, while \(\gammaone\) is legible from the slowest modes.
Against this, the matched benchmark supplies a practice-facing reference:
the tabular learners plateau around \(10^{-4}\) because tree ensembles are piecewise constant and cannot represent an affine map, not because structure was held back.

The controlled design bounds the claims deliberately. The data are synthetic,
the features are noise-free Liouvillian spectra rather than experimentally estimated ones, and no shot noise, calibration drift, or finite-time
measurement protocol enters the pipeline. Coupling topology, disorder, hidden
dissipators, and site-dependent rates are held fixed, and system size varies
only in the scaling study. One structural choice carries most of the weight:
the Hamiltonian of \cref{eq:hamiltonian} commutes with the dephasing
dissipators, so the linear identifiability reported here is a property of that
commuting limit rather than of Liouvillian spectra in general. The
transverse-field scan measures how much that matters --- the equality survives, marginally, to \(h=10^{-2}\) and is not reached by \(h=10^{-1}\) at \(n=4\), bracketed at decade resolution. The paper establishes identifiability from noise-free spectra; it does
not establish a device-ready rate-estimation protocol.

Three qualifications delimit the model comparison. Cross-model ranking is
defined only on the matched grid, since the other families were not run below
\(\featurecount=80\). Selection rests on a finite budget grid and three
overlapping split seeds, so the ordering is descriptive. SHAP and permutation
analyses explain fitted regressors and would require eigenvector tracking
before any coordinate could be promoted to a physical mode identity.

Closing the gap to device characterization is the natural next step: replacing
oracle spectra with finite-shot observables such as decay traces or
reconstructed low-lying poles, varying the Hamiltonian and dissipator
structure, and treating model mismatch and readout error as first-class test
axes. In that regime physics-informed, Bayesian, and symmetry-aware estimators
become the meaningful comparison set.

Spectral identifiability in this commuting Lindblad family has a derivable
resolution threshold. The closed-form spectrum is affine in both dissipative
rates, its \(D=2^n\) population modes carry no dephasing dependence, and the resulting lower bound \(\featurecount\ge D\) is saturated by the measured threshold at each size we test, \(n=4,5,6\). The bound belongs to
the commuting
limit: a transverse field leaves it intact, marginally, to \(h=10^{-2}\), and it is not reached by \(h=10^{-1}\). The result is established for noise-free simulator spectra of a known generator; carrying it to finite-shot, measurement-derived spectra
remains open, and is where the practical value of such a threshold would be
decided.

\section{Methods}

All artifact paths cited in this paper resolve within the archived project
release \cite{ji_2026_22160077} (see Code availability). An automated claims
audit shipped with that release re-checks the reported metrics, thresholds,
and summary statistics of this paper against the archived artifacts. The audit
table records each checked claim together with its source file.

\subsection{Markovian open-system dynamics}
This study assumes time-homogeneous, completely positive, trace-preserving
semigroup dynamics; the density operator \(\rho(t)\in\mathbb{C}^{D\times D}\)
of an \(n\)-qubit Hilbert space with dimension \(D=2^n\) evolves, in units
where the reduced Planck constant is set to \(\hslash=1\), according to the
Lindblad master equation
\begin{equation}
  \frac{\mathrm{d}\rho}{\mathrm{d}t}
  =
  -\mathrm{i}[H,\rho]
  +
  \sum_j
  \left(
    L_j\rho L_j^\dagger
    -
    \frac{1}{2}\{L_j^\dagger L_j,\rho\}
  \right)
  \equiv
  \liouvillian(\rho),
  \label{eq:lindblad}
\end{equation}
where \(H\) is the system Hamiltonian, and each \(L_j\) is a collapse operator
\cite{Gorini1976,Lindblad1976,BreuerPetruccione2002}. Here
\(\{A,B\}=AB+BA\) denotes the anticommutator. For the open quantum system
used in this study, \(n=6\), and
\begin{equation}
  H
  =
  \frac{\omega}{2}\sum_{i=1}^{n}\sigma_i^z
  +
  J\sum_{1\leq i<j\leq n}\sigma_i^z\sigma_j^z,
  \qquad
  \omega=1.0,\quad J=0.05.
  \label{eq:hamiltonian}
\end{equation}
For \(n=6\), the Hilbert-space dimension is \(D=64\).
The parameter \(\omega\) is the single-qubit level splitting, an angular
frequency of the Hamiltonian rather than a fundamental constant; with
\(\hslash=1\) fixed above, setting \(\omega=1\) selects it as the reference
scale of the dimensionless units: time is expressed in units of
\(\omega^{-1}\), while \(J\), \(\gammaone\), \(\gammaphi\), the
transverse-field strength \(h\) introduced below, and
\(\lambda_m\) are expressed in units of \(\omega\).
Here \(\sigma_i^z\) is the Pauli-\(z\) operator acting on qubit \(i\). Two
types of dissipative channels act locally on each qubit:
\begin{equation}
  L_{i,1}
  =
  \sqrt{\gammaone}\,\sigma_i^-,
  \qquad
  L_{i,\phi}
  =
  \sqrt{\gammaphi}\,\sigma_i^z,
  \qquad
  i=1,\ldots,n,
  \label{eq:collapse-ops}
\end{equation}
where \(\sigma_i^-\) is the lowering operator on qubit \(i\). Therefore the sum in
\cref{eq:lindblad} contains \(2n=12\) collapse operators. With the
normalization in \cref{eq:collapse-ops}, \(\gammaphi\) is the coefficient
multiplying the local dephasing dissipator; for an isolated qubit, this term
gives \(\mathrm{d}\rho_{01}/\mathrm{d}t=-2\gammaphi\rho_{01}\). Apart from
the Hamiltonian phase, the two dissipative channels together contribute
\(\mathrm{d}\rho_{01}/\mathrm{d}t=
-(\gammaone/2+2\gammaphi)\rho_{01}\), while amplitude damping gives
\(\mathrm{d}\rho_{ee}/\mathrm{d}t=-\gammaone\rho_{ee}\). Both terms of
\cref{eq:hamiltonian} are diagonal in the computational basis, so \(H\)
commutes with every dephasing operator \(L_{i,\phi}\) and the model sits in the
commuting, purely \(\sigma^z\) limit. This structural choice is held fixed for
the matched benchmark and the system-size scaling study; it is the setting in
which the linear identifiability reported in the Results holds, and it is
deliberately broken by the transverse-field protocol described below. The
dataset-generation script also constructs a six-qubit GHZ density matrix;
however, this matrix is not used in the spectral benchmark, and no time-domain
trajectory is simulated.

\subsection{Liouvillian spectrum}
After vectorization, \(\liouvillian\) is represented by a
\(D^2\times D^2\) matrix. Its right eigenvalue problem is
\begin{equation*}
  \liouvillian\mathbf{v}_m
  =
  \lambda_m\mathbf{v}_m,
  \qquad
  \mathbf{v}_m\in\mathbb{C}^{D^2},
  \quad
  \lambda_m\in\mathbb{C}.
\end{equation*}
The implementation retains eigenvalues satisfying
\(\operatorname{Re}(\lambda_m)<-10^{-10}\), thereby excluding every mode at or
above the threshold, including the steady mode. The retained eigenvalues are
ordered by decreasing real parts, with the signed imaginary part in ascending
order as the secondary key. This secondary key is a deterministic tie-breaker
for numerically equal real parts, not a mode-tracking procedure: no
tolerance-based matching of nearly degenerate modes or conjugate pairs is
performed. Two structural properties of \(\liouvillian\) bear on this. Its
eigenvalues are massively degenerate, depending only on \(n_i+n_j\) and
\(d_H(i,j)\), which follows from the weak symmetries of the generator
\cite{Albert2014Symmetries}; and the vectorized generator is triangular and
hence non-normal, so its eigenvalues are not guaranteed to be well conditioned
under perturbation. The transverse field introduced below lifts the exact population blindness at second order in \(h\). The cutoff removes exactly one mode per sample in every dataset and
at every field value, leaving \(D^2-1\) non-steady modes; the retained mode
closest to the cutoff has \(|\operatorname{Re}\lambda|\) of order
\(\gammaone\), five to nine orders of magnitude above the threshold, so no
physically slow mode is discarded. After sorting, the
spectral input is defined as
\begin{equation}
\begin{aligned}
  a_m &= \operatorname{Re}(\lambda_m), \\
  b_m &= \operatorname{Im}(\lambda_m), \\
  \mathbf{x}_\featurecount
  &= \left(a_1,\ldots,a_\featurecount,
  b_1,\ldots,b_\featurecount\right)^{\mathsf{T}}
  \in \mathbb{R}^{2\featurecount}.
\end{aligned}
\label{eq:features}
\end{equation}
For \(n=6\), at most \(D^2-1=4095\) non-steady modes are available. The
largest budget size, \(\featurecount=400\), therefore corresponds to approximately
\(9.77\%\) of this dimension-count upper bound. An isolated eigenmode
proportional to \(\exp(\lambda_m t)\) has modal e-folding time
\(\tau_m=-1/\operatorname{Re}(\lambda_m)\) when
\(\operatorname{Re}(\lambda_m)<0\); ordering real parts from closest to zero
therefore biases towards the longest modal decay timescales. Actual relaxation of
an initial state or observable also depends on mode amplitudes, eigenvector
conditioning, and overlaps
\cite{BreuerPetruccione2002,Macieszczak2016Metastability,Minganti2018Spectral,MoriShirai2020}.

\subsection{Closed-form spectrum in the commuting limit}
Both terms of \cref{eq:hamiltonian} and every dephasing operator
\(L_{i,\phi}\) are diagonal in the computational basis, while each damping
operator \(\sigma_i^-\) strictly lowers the excitation number. Ordering the
vectorized basis elements \(|i\rangle\langle j|\) by total excitation content
therefore makes the matrix of \(\liouvillian\) triangular, so its eigenvalues
are its diagonal entries, which gives \cref{eq:closed-form} with \(E_i\) the
diagonal energies of \cref{eq:hamiltonian} and \(n_i\) the excitation number
of basis state \(i\). This is an instance of the closed-form class of
Lindblad master equations without gain~\cite{Torres2014ClosedForm} and is not
claimed as novel here. One convention matters for reproduction: QuTiP's
\(\texttt{sigmam()}\) lowers basis index 0 into index 1, so index 0 is the
excited state and \(n_i=n-\operatorname{popcount}(i)\). Reading \(n_i\) as
\(\operatorname{popcount}(i)\) still reproduces the correct eigenvalue
multiset, because bitwise complementation is a bijection that preserves
Hamming distance, but it attaches the wrong imaginary part to individual
entries and fails an entrywise check. The closed form was verified entrywise
against the numerically constructed generator at \(n=4\) over 25 protocol
rate pairs, with maximum absolute deviation \(1.67\times10^{-16}\) and
exactly vanishing upper-triangular residual.

\subsection{Inverse regression task}
Let \(\mathcal{Q}=\{\gammaone,\gammaphi\}\) represent the target set. For sample
\(i\), the target vector and target-wise predictions are
\begin{equation}
\begin{aligned}
  \mathbf{y}_i
  &= \left(\gamma_{1,i},\gamma_{\phi,i}\right)^{\mathsf{T}}, \\
  \widehat{y}_{i,q}
  &= f_{\boldsymbol{\theta}_q}(\mathbf{x}_{i,\featurecount}),
  \qquad q\in\mathcal{Q}, \\
  \widehat{\mathbf{y}}_i
  &= \left(\widehat{y}_{i,\gammaone},
  \widehat{y}_{i,\gammaphi}\right)^{\mathsf{T}}.
\end{aligned}
\label{eq:regression}
\end{equation}
The two dissipative coefficients are sampled independently and uniformly from the following interval:
\([0.005,0.05]\). The following target-wise fitting description applies to
the four matched benchmark families. Each model family fits a single-output regressor for
\(\gammaone\) and another for \(\gammaphi\), after which the two predictions are
combined for target-wise and joint evaluation. This design allows assessment
of whether a model predicts one coefficient more accurately than the other.
The system size, coupling strength, rate interval, sample size, and
feature-budget grids define a controlled synthetic benchmark; they were not
calibrated to a particular physical device.

\subsection{Synthetic dataset generation}
The synthetic datasets were generated with a QuTiP Python script
\cite{Johansson2012QuTiP} that constructs the necessary quantum-mechanical objects. For each sampled pair
\((\gammaone,\gammaphi)\), the code computed the Hamiltonian in
\cref{eq:hamiltonian}, instantiated the collapse operators in
\cref{eq:collapse-ops}, formed the Liouvillian with
\texttt{qt.liouvillian}, and directly computed its eigenvalues. The
Hamiltonian in \cref{eq:hamiltonian} was constructed once and reused across
all sampled pairs. It then
applied the criterion and sorting rule described above and wrote the
first \(\featurecount\) real components followed by the corresponding
\(\featurecount\) imaginary components to a CSV file after casting the spectral
features to 32-bit floating-point values. The two target columns are written
at full double precision and are not quantized; the sub-nanounit regression
errors reported below are therefore limited by the predictor precision alone.
The metadata field contains
\texttt{gamma1}, \texttt{gammaphi}, \texttt{k}, \texttt{n},
\texttt{omega}, and \texttt{J}. No Lindblad master equation trajectory was integrated.
Here, \texttt{gamma1} and \texttt{gammaphi} are target columns, whereas
\texttt{k}, \texttt{n}, \texttt{omega}, and \texttt{J} are provenance fields.
The predictors are ideal Liouvillian eigenvalues computed directly from the
synthetic generator; no measurement protocol, spectral reconstruction,
finite-shot noise, or experimental calibration was modelled.

The five files used for the matched comparison each contain 1000 unique rate
pairs at one of the feature budgets
\(\featurecount\in\{80,160,240,320,400\}\). They use the same ordered rate
pairs and the physical constants \(n=6\), \(\omega=1.0\), and \(J=0.05\). Both
coefficients lie in \([0.005,0.05]\). The saved-file checks are included in the archived validation tables~\cite{ji_2026_22160077}. The saved ordered rate-pair
sequence is reproduced by \texttt{numpy.random.default\_rng} with seed 0.

\subsection{Preprocessing and feature budgets}
In the manuscript and figure labels, only columns beginning with \texttt{eig\_real\_} or
\texttt{eig\_imag\_} are used as predictors. Each suffix is a three-digit,
zero-padded index \(r=0,\ldots,\featurecount-1\), as in
\texttt{eig\_real\_000} and \texttt{eig\_imag\_000}. The corresponding coordinates are denoted \(a_{r+1}\) and \(b_{r+1}\),
respectively, as defined in \cref{eq:features}. The target columns are
\texttt{gamma1} and \texttt{gammaphi}; the provenance fields \texttt{k},
\texttt{n}, \texttt{omega}, and \texttt{J} are excluded from the
regressors. Consequently, the input dimension is \(p=2\featurecount\).

The matched cross-model budget grid is
\(\featurecount\in\{80,160,240,320,400\}\), adjacent to
\(p\in\{160,320,480,640,800\}\). A separate AutoGluon-only study spans \(\featurecount\in\{10,20,30,40,50,60,70\}\) to examine the lower-resolution regime below \(\featurecount=80\). AutoGluon was used for this arm because it recorded the lowest mean joint MAE on the matched grid; as the Results note, that ordering is not resolved at this design's effective sample size, so the arm is a resolution study within one family rather than a statement about which family is best. Because the other three model families
were not evaluated on this grid, the low-\(\featurecount\) sweep is excluded
from cross-model comparisons. Across all 12 retained budgets, the target
pairs occur in the same order and each lower-budget spectrum is an exact
coordinate prefix of the corresponding higher-budget spectrum; these checks are included in the archived validation tables~\cite{ji_2026_22160077}. The reported resolution analysis uses
the 60-evaluation low-budget collection; an earlier four-evaluation collection
is not used.

\subsection{Training, validation, and test protocol}
The tabular benchmark and the simple baselines partition each dataset the same
way for every model family, feature budget and split seed, through two
deterministic calls to
\texttt{train\_test\_split}~\cite{Pedregosa2011Scikit}. The first call set aside
\(20\%\) of the 1000 samples as a test set. The second took the remaining 800
and moved a quarter of them into validation, which left 600 training, 200
validation, and 200 test samples. Both calls received the same random state,
and we repeated the procedure for split seeds 42, 43, and 44. The
real-coordinate threshold scan uses a different partition: a single
\texttt{train\_test\_split} call with \texttt{test\_size}\(=0.2\), giving 800
training and 200 held-out rows, and no validation split, since it fits no
hyperparameters. Unless a split seed is named, every reported metric is the
mean over seeds 42, 43 and 44. Since those seeds
partition one and the same set of 1000 rate pairs, the three splits overlap and
are not independent datasets.

Model selection, together with early stopping wherever a method implemented it,
drew only on the validation partition. The test partition never entered fitting
or hyperparameter search. We computed final metrics from every test partition,
and reused just one of them, XGBoost at \(\featurecount=80\) on split 1, for the
post hoc interpretability diagnostics described below. Within a given
\(\featurecount\) the input CSV stayed fixed across the three splits. Once
selection was complete, the chosen models remained fitted to their 600 training
rows; we did not refit them on the pooled training and validation data.

\subsection{Regression models}
The matched benchmark draws on four families of tabular models. XGBoost,
LightGBM, and CatBoost are gradient-boosted decision-tree methods
\cite{ChenGuestrin2016,Ke2017LightGBM,Prokhorenkova2018CatBoost}, while
AutoGluon is an automated machine-learning framework that assembles and tunes
its own candidate predictors~\cite{Erickson2020AutoGluon}. The four families
were selected from the model-performance ranking of the 111-dataset tabular
benchmark of Shmuel et al.~\cite{Shmuel2025TabularBenchmark}. In each case we
trained one regressor for \(\gammaone\) and a second for \(\gammaphi\), then
paired the outputs for joint scoring. The optimization targets were not
identical across implementations: XGBoost minimized squared error, LightGBM
used its native \(L_1\) criterion, CatBoost trained under root-mean-square-error loss, and
AutoGluon ran in regression mode; all four reported MAE as the evaluation
metric. Three further baselines --- ordinary least squares, ridge regression
and a random forest --- were fitted on the same 600 training rows as a benchmark-arm baseline and
evaluated on the same 200 test rows. The two linear baselines operate on
standardized inputs, with every scaler fitted on the training rows alone.
Because the baseline analysis reads the quantized CSV predictors back as
64-bit arrays while the four matched training scripts cast to 32-bit, widening
the stored values cannot recover precision already lost at write time. This sets the numerical resolution limit of every recovery reported here: the float32 spacing
at a typical retained coordinate \(|\operatorname{Re}\lambda|\approx0.05\) is
\(3.7\times10^{-9}\), and a least-squares fit pools many such coordinates. The
sub-\(10^{-9}\) errors we report for ordinary least squares are of this order
and should be read as recovery limited by the precision of the archived
features rather than as exact algebraic inversion; we claim no specific averaging behaviour for the quantization error, since the retained coordinates are
strongly correlated and frequently degenerate. Full library-level configurations, the ridge penalty search, and the random-forest settings are included in the archived code and analysis release~\cite{ji_2026_22160077}.

\subsection{Additional identifiability and robustness diagnostics}
The diagnostic suite reported in the Results --- sorted-coordinate
trajectories and adjacent-pair degeneracy, real/imaginary feature ablations,
clean-train/noisy-test spectral perturbation, unseen-rate-region holdouts, and
the AutoGluon error maps --- all use the \(\featurecount=80\) dataset and the
same split seeds 42, 43 and 44. The coordinate selections, summary-feature definitions, noise model and seeding, and deterministic holdout masks are included in the archived analysis code~\cite{ji_2026_22160077}.

\subsection{System-size scaling and transverse-field protocols}
The scaling study regenerated the full dataset protocol at \(n=3\), 4, and 5
with the unmodified generation script and identical physics settings
(\(\omega=1.0\), \(J=0.05\), and the same 1000 uniformly sampled rate pairs),
then applied the same real-coordinate OLS threshold scan as at \(n=6\): for
each budget, ordinary least squares is fitted on an intercept and the first
\(\featurecount\) sorted real coordinates, and \(\featurecount^\star\) is the
smallest budget at which the held-out MAE falls below \(10^{-7}\), evaluated
for split seeds 42, 43, and 44. The scanned budgets extend to 24, 40, 72, and
80 coordinates at \(n=3,4,5,6\), respectively. The transverse-field study
adds \(h\sum_i\sigma_i^x\) to \cref{eq:hamiltonian} at \(n=4\) with
\(h\in\{0,10^{-4},10^{-3},10^{-2},10^{-1},5\times10^{-1}\}\) and repeats the same scan up to
\(\featurecount=32\); thresholds not reached by \(\featurecount=32\) are
recorded as right-censored (``\(>32\)'') and never extrapolated. All 1000
samples retain 255 non-steady modes at every \(h\). At \(h=0\) the run
reproduces the closed-form control values
(\(\featurecount^\star(\gammaone)=9\), \(\featurecount^\star(\gammaphi)=16\),
and a dephasing MAE drop of 6.8 orders of magnitude at threshold). Structure
deviation is quantified as the maximum absolute difference between the sorted
non-steady \(\operatorname{Re}\lambda\) multiset at field \(h\) and the
\(h=0\) closed form for the same rate pair, over 25 protocol pairs. The complete protocol and output tables are included in the archived analysis release~\cite{ji_2026_22160077}.

\subsection{Hyperparameter optimization}
For the three gradient-boosting methods we selected configurations by the
arithmetic mean of the two target-wise validation MAEs, running 60
Hyperopt/Tree-structured Parzen Estimator (TPE) evaluations per run~\cite{Bergstra2011TPE} over multi-dimensional
search spaces covering learning rate, tree depth, regularization and sampling
fractions. AutoGluon was given the same 60-evaluation budget, drawn with
replacement from four settings that pair the \texttt{medium\_quality} preset
with the same four estimator seeds, with bagging and stacking disabled. For an
automated framework a configuration is a quality preset rather than a point in
a continuous hyperparameter space, so an equal evaluation count equalizes
budget but not search space, stopping behaviour, or computational work;
AutoGluon's figures are accordingly fixed-preset results. The seed draw is the
minor axis: across the 60 evaluations of a run the validation loss spans at
most \(2.1\%\) in the matched collection and \(0.33\%\) in the
low-\(\featurecount\) collection. The optimization protocol, estimator seeds, selected configurations, exact Hyperopt prior bounds, and runner definitions are provided with the archived code release~\cite{ji_2026_22160077}.

\subsection{Evaluation metrics}
Let \(N=200\) denote the number of held-out test samples in each run, let
\(q\in\mathcal{Q}\) index the targets, and let \(\widehat{y}_{i,q}\) and
\(y_{i,q}\) be the predictions and true coefficients of
\cref{eq:regression}. The
target-wise mean absolute error (MAE), root-mean-square error (RMSE), and
coefficient of determination are
\begin{align*}
  \operatorname{MAE}_q
  &=
  \frac{1}{N}
  \sum_{i=1}^{N}
  \left|\widehat{y}_{i,q}-y_{i,q}\right|,
  \\
  \operatorname{RMSE}_q
  &=
  \left[
    \frac{1}{N}
    \sum_{i=1}^{N}
    \left(\widehat{y}_{i,q}-y_{i,q}\right)^2
  \right]^{1/2},
  \\
  R_q^2
  &=
  1
  -
  \frac{
    \sum_{i=1}^{N}
    \left(\widehat{y}_{i,q}-y_{i,q}\right)^2
  }{
    \sum_{i=1}^{N}
    \left(y_{i,q}-\overline{y}_q\right)^2
  },
  \qquad
  \overline{y}_q
  =
  \frac{1}{N}
  \sum_{i=1}^{N}y_{i,q}.
\end{align*}
Joint MAE is the arithmetic mean of the two target-wise MAEs,
\begin{equation*}
\begin{aligned}
  \operatorname{MAE}_{\mathrm{joint}}
  &= \frac{1}{2}\left(
  \operatorname{MAE}_{\gammaone}+\operatorname{MAE}_{\gammaphi}\right) \\
  &= \frac{1}{2N}
  \sum_{q\in\mathcal{Q}}\sum_{i=1}^{N}
  \left|\widehat{y}_{i,q}-y_{i,q}\right|.
\end{aligned}
\end{equation*}
Joint RMSE is instead computed by pooling all \(2N\) target--sample errors,
\begin{equation*}
\begin{aligned}
  \operatorname{RMSE}_{\mathrm{joint}}
  &= \left[
  \frac{1}{2N}\sum_{q\in\mathcal{Q}}\sum_{i=1}^{N}
  \left(\widehat{y}_{i,q}-y_{i,q}\right)^2
  \right]^{1/2} \\
  &= \left[
  \frac{\operatorname{RMSE}_{\gammaone}^{2}
  +\operatorname{RMSE}_{\gammaphi}^{2}}{2}
  \right]^{1/2}.
\end{aligned}
\end{equation*}
Both targets received equal weight because they have the same units and were
sampled over the same interval. The square-root operation is applied only
after pooling the squared errors for \(\operatorname{RMSE}_{\mathrm{joint}}\);
it is not applied to \(\operatorname{MAE}_{\mathrm{joint}}\), and the pooled
joint RMSE is not the arithmetic mean of the two target-wise RMSEs.
The prediction files also store absolute and squared errors for each target,
enabling residual, quantile, and worst-case analyses.

\subsection{Statistical analysis}
Every model comparison rests on \(n_b=15\) matched
\((\featurecount,\mathrm{split})\) conditions, with \(d_b\) the paired
difference in joint MAE between two models in block \(b\). For each of the six
model pairs we tested the 15 differences for normality with a Shapiro--Wilk
test~\cite{ShapiroWilk1965} and applied a paired \(t\)-test or a two-sided
Wilcoxon signed-rank test~\cite{Wilcoxon1945} accordingly, correcting the six
\(p\)-values by the Holm and Benjamini--Hochberg procedures
\cite{Holm1979,BenjaminiHochberg1995}. Effect sizes are Cohen's \(d_z\) and
Cliff's delta~\cite{Cliff1993}; interval statements use percentile intervals
from \(10{,}000\) bootstrap resamples seeded at 42~\cite{Efron1979}.
These steps reproduce the saved analysis faithfully, but their standing is
limited: the feature budgets are nested views of the same rate pairs and the
repeated splits overlap, so the 15 block values are correlated rather than
independent, and a normality pretest does not repair that dependence. We
therefore read the \(p\)-values and intervals as exploratory descriptions of
these particular runs rather than confirmatory inference; genuine confirmatory
tests would require independently generated datasets
\cite{NadeauBengio2003,SchucanyNg2006}. The complete test definitions, boundary-test results, and residual-normality protocol are included in the archived analysis release~\cite{ji_2026_22160077}.

\subsection{Interpretability analysis}
Interpretability is treated as a post-hoc analysis of one fitted regressor,
not as evidence of causality. TreeSHAP contributions
\cite{LundbergLee2017,Lundberg2020TreeSHAP}, a ten-repeat permutation-importance
check~\cite{Breiman2001RandomForests}, and clustered Spearman correlations were
computed for the saved XGBoost \(\featurecount=80\) split-1 models on the
200-row test partition. Because the spectrum is sorted independently for each
sample, the resulting coordinates are operational feature indices rather than
tracked physical eigenmodes. The full interpretability protocol is included in the archived analysis release~\cite{ji_2026_22160077}.

\subsection{Software and computational environment}
The saved matched-run artifacts record XGBoost 3.2.0, LightGBM 4.6.0,
CatBoost 1.2.10, and AutoGluon 1.5.0. The matched runners were invoked in CUDA
mode; individual AutoGluon constituent models could nevertheless execute on
the CPU. Where captured, the remaining Python, QuTiP, NumPy, SciPy,
scikit-learn, Hyperopt, SHAP, operating-system, and hardware versions are
recorded in the environment manifest (\texttt{ENVIRONMENT.md}) accompanying
the archived code release, which separates recorded values from
reconstructed ones and lists residual gaps explicitly.
\section{Data availability}
The datasets generated and analysed during the current study, including the
simulated Liouvillian spectra, derived identifiability metrics, and numerical
results underlying the figures and tables, are publicly archived on Zenodo
\cite{ji_2026_22160077} under DOI
10.5281/zenodo.22160077. Summary results and additional diagnostics supporting the findings are included in the archived release.

\section{Code availability}
All custom code used to generate the simulations, construct the spectral
datasets, fit the models, perform the statistical analyses, and reproduce the
figures is publicly archived on Zenodo
\cite{ji_2026_22160077} under DOI
10.5281/zenodo.22160077. The archived release includes the automated claims
audit used to re-check the reported metrics, thresholds, and summary statistics
against the saved artifacts. Component versions are recorded in the environment
manifest accompanying the archive.

\section{Use of large language models}
OpenAI Codex, Anthropic Claude Fable 5 and Anthropic Claude Opus 5 were used
under the authors' direction in two roles: editing the manuscript for grammar
and readability, and writing the simulation and analysis code with which the
reported experiments were run. They were also used for \LaTeX{} formatting and
for consistency checks of the source against the saved artifacts. They were
not used to generate or alter any reported quantity: every number in this
paper is produced by the archived code from the archived data, and re-checked
against those artifacts by the automated claims audit. The authors reviewed
the outputs and remain accountable for the final manuscript, analysis, and
submission files.

\section{Author contributions}
Y.J. conceived the study, generated the benchmark data, performed the analysis,
prepared the figures, and wrote the manuscript.
S.C. provided technical guidance on the machine-learning methodology,
advising on the selection of model families and on the design of the
benchmarking protocol. All authors reviewed and approved the final
manuscript.

\section{Funding}
This work received no specific grant from any funding agency in the public,
commercial, or not-for-profit sectors.

\section{Competing interests}
The authors declare no competing interests.

\begin{acknowledgments}
The authors acknowledge the open-source scientific Python ecosystem, whose
libraries are cited individually in the text.
\end{acknowledgments}

\bibliography{references}

\end{document}